# Robust Prediction Model for Multidimensional and Unbalanced Datasets

*Pooja Thakar[a], Anil Mehta[b], Manisha[a]\**

[a]*Banasthali University, Jaipur, Rajasthan 304022, India*
[b]*Manipal University, Jaipur, Rajasthan 303007, India*

Abstract:

Data Mining is a promising field and is applied in multiple domains for its predictive capabilities. Data in the real world cannot be readily used for data mining as it suffers from the problems of multidimensionality, unbalance and missing values. It is difficult to use its predictive capabilities by novice users. It is difficult for a beginner to find the relevant set of attributes from a large pool of data available. The paper presents a Robust Prediction Model that finds a relevant set of attributes; resolves the problems of unbalanced and multidimensional real-life datasets and helps in finding patterns for informed decision making. Model is tested upon five different datasets in the domain of Health Sector, Education, Business and Fraud Detection. The results showcase the robust behaviour of the model and its applicability in various domains.

## 1. Introduction

The data mining is a promising field and is growing in popularity due to its vast applicability in various areas. But it has to face many challenges before it can be put to use for resolving real-life problems. Real life data is generally not fit for direct data mining and suffers from many problems. Data is generally large, multidimensional and unbalanced in nature with lots of missing values. Plenty of publically available data mining tools are available, but they can be used only when data fed in them have the capability to produce relevant information. Data should be treated well before feeding in any tool. Entrepreneurs, Managers and new setup businesses may not always be well versed with data mining intricacies and such a powerful tool cannot be used easily for decision making. To make novice users adapt and apply modern techniques and practices, a robust prediction model is proposed in this paper. In real-life datasets, data is generally unbalanced. Data is unbalanced when some classes have a significantly large number of instances available than others. Predictive algorithms overlook minority classes and consider only majority classes. Minority class hubs may lead to misclassification in many high-dimensional datasets (Tomašev, 2013). As a result, the predictive algorithms are unable to classify them correctly. A controlled dataset with high dimensionality may result in over-fitting and decrease the generalization performance of predictive algorithms (Tachibana, 2014). This, in turn, requires a huge set of data for better prediction. Furthermore, all the attributes in dataset don't contribute to the final prediction. Hence, finding the relevant set of attributes from a large pool of attributes is also a challenging task. To overcome the aforesaid problems a Robust Prediction Model (RPM) is designed which can be used with varied datasets that are multidimensional, large and unbalanced in nature. Robust Prediction Model works on two major aspects.

- To find the major set of attributes in the data-set that affects the prediction in an automated approach.
- To predict target class by integrating machine learning methods of classification for better decision making.

The RPM model is robust in nature. This paper applies the RPM model on five different types of datasets from the domains of medicine, marketing, education and fraud detection. Results prove the robustness and applicability of Robust Prediction Model in various fields. RPM is applied to public datasets from the UCI Repository (Dheeru, 2017). The datasets are as follows:

- "Epileptic Seizure Recognition Data Set", which belongs to the health sector. Every data is EEG recording value at a different point in time (Andrzejak, 2001).
- "Student Performance Data Set", which belongs to the education sector and predicts student achievement in secondary education of two schools in Portuguese. Two datasets on Mathematics and Portuguese Language are provided (Silva, April 2008).
- "Turkiye Student Evaluation Dataset", which also belongs to the education sector. Dataset is composed of "evaluation scores" provided by students. Students are from "Gazi University" in Ankara, Turkey (Gunduz, 2013).
- "Bank Marketing Dataset", which belongs to the marketing sector in business management. Data is composed of results of direct marketing campaigns through phone calls for a Portuguese bank. The target is to predict "if the client will subscribe to a term deposit" (Moro, 2014).
- "Polish Companies bankruptcy data dataset", which belongs to fraud detection. The dataset is composed of Polish companies' bankruptcy prediction. The bankrupt companies were analyzed in the period 2000-2012. "Still operating companies" were evaluated from 2007-2013 (Zięba, 2016).

RPM resulted in very encouraging results with high accuracy level. The user doesn't need to understand intricate details of dimensionality reduction and prediction yet can apply the model for informed decision making.

Paper further describes Robust Prediction Model (RPM) in detail in Section 2, Section 3 describes the type of datasets on which RPM is applied for the test of robustness. Section 4 depicts the results obtained. It also compares the results of RPM with PCA (Principle Component Analysis) method of dimensionality reduction and Section 5 concludes the research with its future scope.

## 2. Robust Prediction Model (RPM)

A Robust Prediction Model (RPM) has been designed by integrating unsupervised and supervised learning techniques of clustering and classification (Thakar, Mehta and Manisha, 2017) (Hu, 2017). It helps in predicting the target class of the dataset. RPM model works in three phases. In the first phase, the concept of automated pre-processing is performed, where raw data is converted to a ready dataset that helps in better prediction through classification (Thakar, Mehta, & Manisha, 2016). It reduces dimensionality and finds a relevant set of attributes, which can further be readily used for better classification results. The" k-means clustering" is applied to attribute set for the purpose of finding a relevant set of attributes (Kim, 2005). In the second phase, dataset received from the first phase undergoes ensemble vote classification. Instead of choosing one method for classification, voting ensemble method is used for improved classification accuracy. The third phase finds rules from results obtained in the second phase. Robust Prediction Model (RPM) thus deals with the complex dataset, automatically selects a relevant set of attributes from a large pool of attributes and integrates learning algorithms to predict target class. Model is scalable enough to be applied in any domain. Thus, proposes an easy and generalized solution to prediction problems in real life. The details on all three phases of the model are described below:

Phase I: At this stage, the dataset is pre-processed in an innovative way, where the raw dataset is fed into the system to find a relevant set of attributes and transform the dataset into refined data that can be readily used for classification. The raw data set is first balanced by Sample Bootstrapping and equal instances of each class are taken into consideration. Thereafter, k- means clustering is applied on the balanced dataset after transposing attributes to instances and instances to attributes. It results in sets of clusters (clusters of attributes), where the related sets of attributes are clubbed together in one cluster. All obtained clusters are re-transposed and tested for its capabilities. Simple Cart classification is applied to all clusters (Denison, 1998). Cluster with better classification accuracy is then selected and taken further to the next level. At the next level it is transposed again, k- means clustering is re-applied on the chosen cluster. This process can be repeated n times unless the chosen cluster produces better results. After n level clustering, clusters are obtained by the process of filtration. Finally, transpose clusters and apply Chi-Square to select top attributes and obtain refined and transformed dataset with selected attributes. This automated approach helps in fast and easy selection of relevant attributes from a large pool of attributes and also enhance the quality of dataset for further classification. It results in improved classification accuracy (Thakar, Mehta and Manisha, 2017).

Phase II: At second phase transformed dataset derived from 'Phase I' is used for classification. Instead of choosing one method for classification, voting ensemble method is used for improved classification results. Vote method uses the vote of each base learner for classification of an instance; the prediction with maximum votes is taken into consideration. It uses predictions of base learners to make a combined prediction by using majority voting method.

Phase III: Results are obtained from 'Phase II' and they become the basis to generate rules. Results produced by classification methods help in writing rules for the dataset.

Abbreviations for Algorithm of RPM:

| | |
|---|---|
| $D_{un}$: | Variable to store the unprocessed/raw dataset |
| SB(): | Method for Sample Bootstrapping |
| $CL_1..CL_m$: | Depicts target classes of the raw dataset, where m is the total number of target classes |
| $D_1$: | Sampled dataset after combining each target class |
| Append(): | Method for appending instances |
| TRAN(): | Method to transpose matrix |
| $C_x$: | Selected Cluster |
| $D_2$: | Transposed dataset/cluster |
| $DC_1..DC_n$: | Depicts dataset clusters after separating clusters, where n is the total number of clusters |
| $C_1..C_n$: | Re-transposed clusters, where n is the total number of clusters |
| $D_f$: | Final dataset for classification |
| $P_i$: | Prediction Accuracy of the dataset, where i represents the number of iteration |
| MAX(): | To find the cluster with maximum prediction accuracy |
| CHI(): | Method to find chi-square weights of the cluster |
| SELECT(): | Select v attributes from dataset/cluster, where v is number of attributes |

The Algorithm of RPM Model:

Step1: Load Raw/ Unprocessed Dataset ($D_{un}$) in the form of a matrix and initialise Cluster $C_x$ to Null and i=1.

Step2: Perform Sample Bootstrapping SB () on $D_{un}$ i.e. SB ($D_{un}$, m) w.r.t. a number of classes (m) in $D_{un}$, where, $D_{un} \in CL_1..CL_m$

Step3: Select equal samples of each class $CL_1..CL_m$ and create dataset $D_1$ i.e. $D_1$=Append (SB ($D_{un}$,m))

Step4: Transpose TRAN () matrix received ($D_1$ or $C_x$) and create dataset $D_2$ i.e. $D_2$= TRAN ($D_1$) or TRAN ($C_x$)

Step5: Apply k-means clustering on $D_2$ where k=number of classes in $D_2$

Step6: Filter clusters of $D_2$ w.r.t. cluster number and generate new datasets as Data Clusters $DC_1$, $DC_2$..$DC_n$

Step7: Transpose all Data Clusters as
$C_1$=TRAN ($DC_1$)
$C_2$=TRAN ($DC_2$)
$C_n$=TRAN ($DC_n$)

Step8: Implement Simple Cart Classification on each Data Clusters $C_1$..$C_n$

Step9: Validate performance of all clusters and find a cluster with maximum prediction accuracy i.e. $C_x$=MAX ($C_1$,$C_2$..$C_n$)

Step11: Apply Chi-square CHI ($C_x$) on the received cluster and find weights of each attribute.

Step12: Select top v attributes with maximum weights from the selected cluster and create final dataset $D_f$ i.e. $D_f$=SELECT (CHI ($C_x$), v))

Step13: Apply ensemble vote classification on $D_f$ and find performance $P_i$ with 10 cross-validations

Step14: If $P_i < P_{i-1}$ or i==1, then i=i+1 and GOTO Step 4 with $C_x$, else GOTO Step 15

Step15: Generate rules with results obtained, where $P_i < P_{i-1}$

In the process of automation, data is loaded into the system. Unprocessed dataset ($D_{un}$) go through Sample Bootstrapping SB ($D_{un}$,m) w.r.t. a number of classes (m) in $D_{un}$. This produces balanced dataset ($D_1$) by applying append function (Append (SB ($D_{un}$,m)). For the purpose of balancing, equal instances of each class are taken into consideration. Now, the balanced dataset ($D_1$) is transposed by transposing attributes to instances and vice versa that converts dataset $D_1$ to $D_2$. Thereafter, k-means clustering is applied to the transposed dataset $D_2$, where k is the number of classes in $D_2$. This produces n sets (n is set as per the number of target classes in the dataset) of clusters (clusters of attributes). Each cluster clubs together the related set of attributes. All the clusters are filtered separately to produce clustered datasets $DC_1$ to $DC_n$. All the clusters are re-transposed for classification thus produces clusters $C_1$ to $C_n$. Simple Cart classification is then applied to all the clusters i.e. $C_1$ to $C_n$. Performance of all the clusters from $C_1$ to $C_n$ is tested with 10 cross-validations. Cluster with best classification result is selected ($C_x$, where $C_x$=MAX($C_1$ to $C_n$)). Chi-square is applied to the selected cluster. Top v attributes are selected to produce dataset $D_f$, where $D_f$ = SELECT (CHI ($C_j$), v). Ensemble vote is applied to $D_f$ and Performance $P_i$ is noted. Here $C_x$ is taken further to the next level and other clusters are simply discarded. At next level clustering, the chosen cluster ($C_x$) is transposed again. It undergoes clustering again with k-means clustering and whole the process is repeated to produce Performance $P_i$. At this stage $P_i$ is compared with $P_{i-1}$ and decision is taken where further clustering of the obtained cluster is required or not. If $P_i < P_{i-1}$, Then cluster is taken further for iteration otherwise $P_{i-1}$ is used for generating rules.

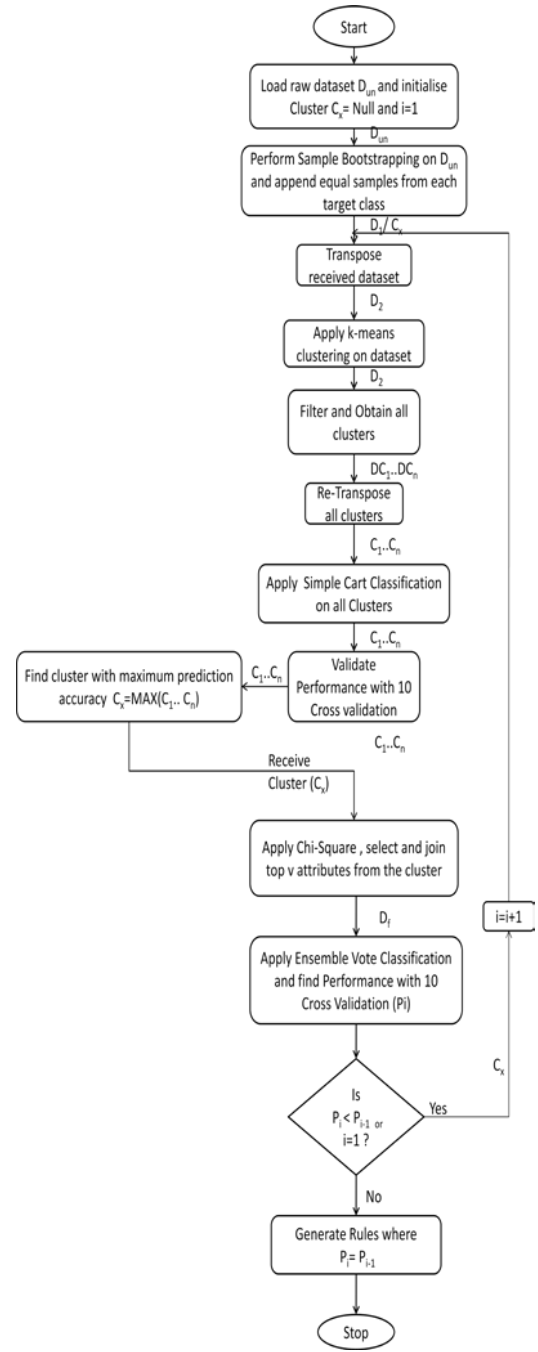

**Fig 1: Logical Diagram of RPM**

This automated pre-processing approach helps in parsimonious selection of relevant attributes from a large pool of attributes in a fast and easy way (Thakar, Mehta and Manisha, 2018), (Thakar, Mehta and Manisha, 2015). This enhances the quality of dataset for classification and results in improved classification accuracy. Further, classification is performed on transformed dataset derived from the previous phase. The voting ensemble method is used to improve classification accuracy. Instead of choosing one method for classification, an ensemble method is used (Catal & Nangir, 2017). Base learners are selected as they perform better than adjoining classifiers when applied alone. Vote method of ensemble classification employs the vote of each learner for classification of an instance; the prediction with maximum votes is taken into consideration. Base learners' predictions are combined to make a final prediction by using majority voting method. Rules thus generated can be readily used for the purpose of decision making. The user only has to feed in the raw dataset. The model will produce relevant rules and patterns that can be used for informed decision making. To test the model it is applied to various datasets available publically in different domains but with the common problem of multidimensionality, unbalance and large dataset with missing values.

## 3. Application of Robust Prediction Model

### 3.1. Datasets

To test the robustness of RPM, it is applied to various public datasets. All the datasets were taken from "The UCI Machine Learning Repository" (Dheeru, 2017). The descriptions of all datasets are as follows:

DATASET A: "Epileptic Seizure Recognition Data Set" records EEG recording at a different point in time (Andrzejak, 2001). Dataset is pre-processed and restructured version of commonly used dataset featuring epileptic seizure detection. All cases that fall in classes 2, 3, 4, and 5 are subjects, who did not have an epileptic seizure. Cases in class 1 had an epileptic seizure. Dataset was pre-processed to convert it into a binary class dataset with Class 1 as epileptic seizure and rest as Class 0 representing the absence of an epileptic seizure. Data is large, multidimensional and unbalanced with very fewer instances belonging to the class of subjects having an epileptic seizure. The goal is to predict the presence and absence of epileptic seizures in subjects.

DATASET B: "Student Performance Data Set", predicts student achievement in secondary education of two schools in Portuguese. Two datasets on Mathematics and Portuguese Language are provided (Silva, April 2008). The goal is to predict student performance in secondary education in two subjects (Mathematics and Portuguese) separately. The class G3 has a strong correlation with G2 and G1 attributes because G3 is the final year grade, which was issued at the third period, while G1 and G2 correspond to the first and second period grades. Predicting G3 without G2 and G1 is difficult, but such prediction is much more useful (Silva, April 2008). The model was applied to predict G3 without G1 and G2. Data were converted into binary (Pass and Fail) class in both the datasets, PASS value if G3>=10 and FAIL if G3<10.

DATASET C: "Turkiye Student Evaluation Dataset", the dataset is composed of a total 5820 evaluation scores submitted by students from "Gazi University" in Ankara, Turkey (Gunduz, 2013) (Oyedotun, 2015). There were 28-course specific questions asked with 5 additional attributes. Q1-Q28 were all Likert-type ranging {1,2,3,4,5}. The target attribute "nb.repeat" produces 3 outputs 1,2,3 that indicate that how many times, of course, will be repeated by the student.

DATASET D: "Bank Marketing Dataset", is a result of direct marketing campaigns through phone calls of a Portuguese banking institution. The classification goal is to predict if the client will subscribe to a term deposit (Moro, 2014). The target class is binary with "YES" or "NO" as per the response of client.

DATASET E: "Polish Companies bankruptcy data dataset", for bankruptcy prediction of Polish companies. The companies that were bankrupt were analyzed in the period 2000-2012, while the still operating companies were evaluated from 2007-2013 (Zięba, 2016). The data was collected from "Emerging Markets Information Service". Based on the collected data 5 classification cases were distinguished depending on the forecasting period. Out of 5, the model is tested on 3 datasets. They are 1st Year data, the data contains financial rates from 1st year of the forecasting period and corresponding class label that indicates bankruptcy status after 5 years. The data contains 7027 instances "financial statements", 271 represents bankrupted companies, 6756 firms that did not bankrupt in the forecasting period. 2nd Year data, the data contains financial rates from the 2nd year of the forecasting period and corresponding class label that indicates bankruptcy status after 4 years. The data contains 10173 instances "financial statements", 400 represents bankrupted companies, 9773 firms that did not bankrupt in the forecasting period. 5th Year data, the data contains financial rates from the 5th year of the forecasting period and corresponding class label that indicates bankruptcy status after 1 year. The data contains 5910 instances "financial statements", 410 represents bankrupted companies, 5500 firms that did not bankrupt in the forecasting period.

### 3.2. Experimental Setup and Measures

RapidMiner Studio Educational Version 7.6.003 was used to implement RPM. The version also extends and implements algorithms designed for Weka Mining Tool. The "10-fold cross-validation" is used as an estimation approach to finding classifier performance. Since there is no separate test data set, thus, this technique divide training dataset into ten equal parts, out of which nine are applied as a training set for making machine learning algorithm learn and one part, is used as a test dataset. This approach is enforced ten times on the same dataset, where every training set has to act as test set once. The number of correct predictions made, divided by the total number of predictions made and multiplied by 100 to turn it into a percentage is known as classification accuracy. In a real-life problem, where there is a large class imbalance. The accuracy paradox may exist, where a model can predict the value of the majority class for all predictions and achieve high classification accuracy (Chawla, 2009). The datasets used in the study suffers from this; hence accuracy measure may not be the only perfect indicator to judge the performance.

Classification accuracy is not enough to make a decision on the effectiveness of the model. Weighted Mean Precision (WMR) and Weighted Mean Recall (WMR) takes False Positive and False Negative into account and is used as another measure for testing performance (Kumar & Rathee, 2011), (Sokolova, 2009). The benefit of these metrics is that they aggregate precision and recall over the result set (Elmishali et al., 2016), (Ori BarIlan et al.). Kappa is another measure, which is used in this case. Kappa Statistics is a normalized value of agreement for a chance (Witten, 2016).

### 3.3. Application of RPM on all Datasets (A-E)

RPM was applied to each dataset separately. The model works at three levels. The first level implements the concept of automated pre-processing, where raw dataset was converted to refined data. Then it was taken further at the second level for classification. An ensemble method of voting with best classifiers for the respective case was used. Last third level generates rules to facilitate decision making. Principle Component Analysis is a well-known method of dimensionality reduction and is widely used to find the relevant set of attributes from the dataset, To overcome the problem of multidimensional data, PCA is considered as very versatile, oldest and the most popular technique in multivariate analysis (Abdi, Hervé, and Lynne, 2010). To test the RPM model, comparative analysis is done on all datasets (A-E). The results are obtained by applying PCA to the same dataset. Instead of applying automated preprocessing of RPM to find the relevant set of attributes, PCA was applied in Phase 1 of RPM. PCA now finds the relevant set of attributes. Further, the ensemble vote method of classification is applied, just as in phase 2 of RPM, which is followed by phase 3 of RPM to find the rules. Thus, only phase 1 of RPM is replaced with PCA. The results clearly showcase improved performance when RPM is applied with all its phases.

## 4. Results

RPM was applied to varied datasets (Dataset A to E) and results were calculated. Results obtained after applying RPM are depicted in terms of Accuracy, Kappa Statistics, Weighted Mean Recall and Weighted Mean Precision.

### 4.1. Results of DATASET A

The model was applied to Epileptic Seizure Recognition dataset and results are shown in Table 1. After sampling, numbers of instances in the dataset were 10,000. RPM model identifies top 12 significant attributes with the process of the automated pre-processing capability of the model and results in 95.2% accuracy of predicting Epileptic Seizure in the subjects.

**Table 1: Results of Dataset A with RPM**

| Parameter | Result |
| --- | --- |
| Accuracy: | 95.21% +/- 0.66% |
| Kappa : | 0.904 +/- 0.013 |
| Weighted Mean Recall: | 95.21% +/- 0.66% |
| Weighted Mean Precision: | 95.31% +/- 0.68% |

The data suffered from two intrinsic problems of multidimensionality and unbalanced dataset. RPM could overcome the problems and results are excellent with 95.21% of accuracy and 0.904 Kappa statistics. This proves the robustness capability of the model to deal with any type of dataset that are large, multidimensional and unbalanced in nature. Ensemble Vote Method combined four base classifiers namely Random Tree, kStar, Simple Cart and Random Forest (Breiman, 2001). Results were obtained after applying PCA on the same dataset followed by RPM phase 2 and 3. Results are shown below in Table 2. The results are compared with RPM results and it is found that model (RPM) outperforms PCA. The graph in Fig. 2 depicts the comparative results.

**Table 2: Results of Dataset A with PCA**

| Parameter | Result |
| --- | --- |
| Accuracy: | 80.00% +/- 0.00% |
| Kappa : | 0.000 +/- 0.000 |
| Weighted Mean Recall: | 50.00% +/- 0.00% |
| Weighted Mean Precision: | 40.00% +/- 0.00% |

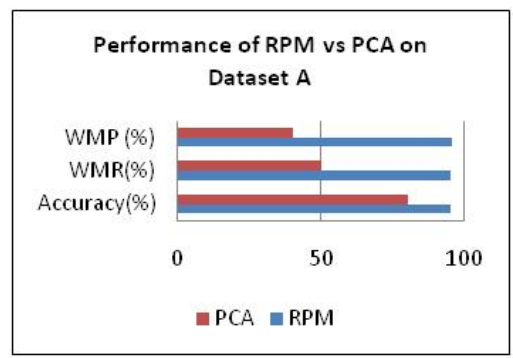

**Fig. 2: RPM and PCA for DATASET A**

### 4.2. Results of DATASET B

RPM Model was applied on "Student Performance Data Set" and results are shown in Table 3. After sampling, numbers of instances in the dataset were 600 and 400 for Portuguese and Mathematics respectively. The class attribute G3 has a strong correlation with G1 and G2. This occurs because G3 is the final year grade that was issued in the third period, while G1 and G2 correspond to the first and second period grades. It is more difficult to predict G3 without G2 and G1. The model was applied to predict G3 without G1 and G2. Data were converted into binary (Pass and Fail) class in both the datasets, PASS value if G3>=10 and FAIL if G3<10. RPM model identifies top 12 significant attributes with the process of the automated pre-processing capability of the model and results in 91.67% and 84.25% accuracy of predicting academic performance in Portuguese and Mathematics respectively. Ensemble Vote Method combined four base classifiers namely Random Tree (Aldous, 1991), kStar, LMT and Random Forest in both the dataset. Results obtained after applying PCA on the same dataset are shown below in fig 3 and 4:

**Table 3: Results of dataset B (Portuguese) with RPM**

| Parameter | Result |
| --- | --- |
| Accuracy: | 91.67% +/- 3.87% |
| Kappa : | 0.833 +/- 0.077 |
| Weighted Mean Recall: | 91.67% +/- 3.87% |
| Weighted Mean Precision: | 92.22% +/- 3.40% |

**Table 4: Results of dataset B (Mathematics) with RPM**

| Parameter | Result |
| --- | --- |
| Accuracy: | 84.25% +/- 3.88% |
| Kappa : | 0.685 +/- 0.078 |
| Weighted Mean Recall: | 84.25% +/- 3.88% |
| Weighted Mean Precision: | 84.90% +/- 3.70% |

**Table 5: Results of dataset B (Portuguese) with PCA**

| Parameter | Result |
| --- | --- |
| Accuracy: | 84.29% +/- 1.12% |
| Kappa : | 0.020 +/- 0.070 |
| Weighted Mean Recall: | 50.64% +/- 2.22% |
| Weighted Mean Precision: | 52.38% +/- 20.27% |

**Table 6: Results of dataset B (Mathematics) with PCA**

| Parameter | Result |
| --- | --- |
| Accuracy: | 62.54% +/- 4.82% |
| Kappa : | 0.114 +/- 0.095 |
| Weighted Mean Recall: | 55.43% +/- 4.39% |
| Weighted Mean Precision: | 56.45% +/- 5.67% |

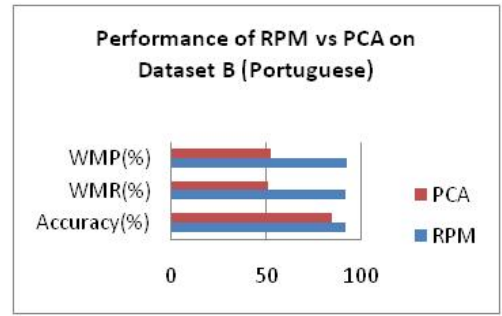

**Fig. 3: RPM and PCA for DATASET B (Portuguese)**

Results showcase that performance of RPM is better than PCA.

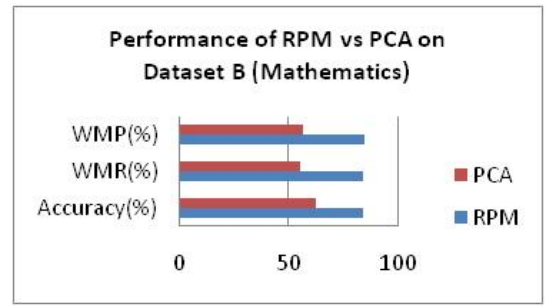

**Fig. 4: RPM and PCA for DATASET B (Mathematics)**

*Results of DATASET C*

The model was applied to "Turkiye Student Evaluation Dataset" and results are shown in Table 7. The instances after sampling become 6000. RPM model identifies top 12 significant attributes with the process of the automated pre-processing capability of the model and results in 83.3% accuracy of predicting the number of times a student repeats a course (1,2 or 3). Ensemble Vote Method combined two base classifiers namely Random Tree and ibk. Comparative Analysis on Dataset C with RPM and PCA is showcased in Fig 5.

**Table 7: Results of dataset C with RPM**

| Parameter | Result |
| --- | --- |
| Accuracy: | 83.30% +/- 1.20% |
| Kappa : | 0.749 +/- 0.018 |
| Weighted Mean Recall: | 83.30% +/- 1.20% |
| Weighted Mean Precision: | 83.65% +/- 1.22% |

**Table 8: Results of Dataset C with PCA**

| Parameter | Result |
|---|---|
| Accuracy: | 82.11% +/- 0.57% |
| Kappa : | 0.102 +/- 0.028 |
| Weighted Mean Recall: | 37.34% +/- 1.47% |
| Weighted Mean Precision: | 42.77% +/- 3.06% |

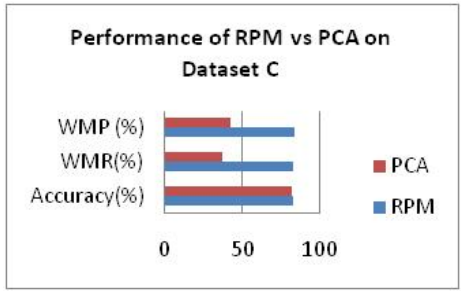

**Fig. 5: RPM and PCA for DATASET C**

*4.3. Results of DATASET D*

The model was applied to "Bank Marketing" and results are shown in Table 9. RPM model identifies top 6 significant attributes with the process of the automated pre-processing capability of the model and results in a 91.75% accuracy of predicting if the client will subscribe to a term deposit. The number of instances after sampling becomes 30,000. Ensemble Vote Method combined two base classifiers namely kstar (Hernández, 2015) and ibk. Comparative Results with PCA are showcased in Fig 6.

**Table 9: Results of dataset D with RPM**

| Parameter | Result |
|---|---|
| Accuracy: | 91.75% +/- 0.40% |
| Kappa : | 0.835 +/- 0.008 |
| Weighted Mean Recall: | 91.75% +/- 0.40% |
| Weighted Mean Precision: | 92.16% +/- 0.41% |

**Table 10: Results of Dataset D with PCA**

| Parameter | Result |
|---|---|
| Accuracy: | 85.42% +/- 0.41% |
| Kappa : | 0.106 +/- 0.014 |
| Weighted Mean Recall: | 54.20% +/- 0.59% |
| Weighted Mean Precision: | 57.70% +/- 1.07% |

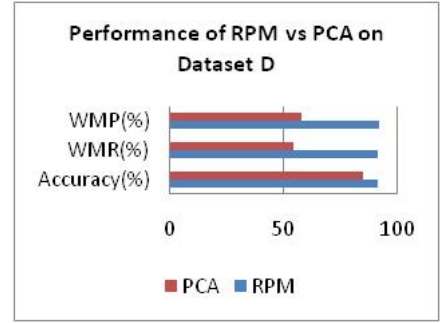

**Fig. 6: RPM and PCA for DATASET D**

*4.4. Results of DATASET E*

The model was applied to "Polish Companies bankruptcy data dataset" for bankruptcy 1, 2 and 5 years and results are shown in Table 11 to Table 13. RPM model identifies top 12 significant attributes with the process of the automated pre-processing capability of the model and results in 93.12%, 87.49% and 90.94% accuracy for bankruptcy dataset 1,2 and 5 respectively of predicting if the company will bankrupt in forecasting period.

**Table 11: Results of dataset E (Bankruptcy 1) with RPM**

| Parameter | Result |
|---|---|
| Accuracy: | 93.12% +/- 0.48% |
| Kappa : | 0.862 +/- 0.010 |
| Weighted Mean Recall: | 93.12% +/- 0.48% |
| Weighted Mean Precision: | 93.96% +/- 0.37% |

**Table 12: Results of dataset E (Bankruptcy 2) with RPM**

| Parameter | Result |
|---|---|
| Accuracy: | 87.49% +/- 1.02% |
| Kappa : | 0.750 +/- 0.020 |
| Weighted Mean Recall: | 87.49% +/- 1.02% |
| Weighted Mean Precision: | 90.00% +/- 0.65% |

**Table 13: Results of dataset E (Bankruptcy 5) with RPM**

| Parameter | Result |
|---|---|
| Accuracy: | 90.94% +/- 1.25% |
| Kappa : | 0.819 +/- 0.025 |
| Weighted Mean Recall: | 90.94% +/- 1.25% |
| Weighted Mean Precision: | 92.35% +/- 0.90% |

Ensemble Vote Method combined two base classifiers namely kstar and ibk. Results were obtained after applying PCA and are depicted in Fig 7 to 9.

**Table 14: Results of Dataset E (Bankrupcy1) with PCA**

| Parameter | Result |
|---|---|
| Accuracy: | 96.11% +/- 0.13% |
| Kappa : | -0.001 +/- 0.001 |
| Weighted Mean Recall: | 49.99% +/- 0.03% |
| Weighted Mean Precision: | 48.07% +/- 0.05% |

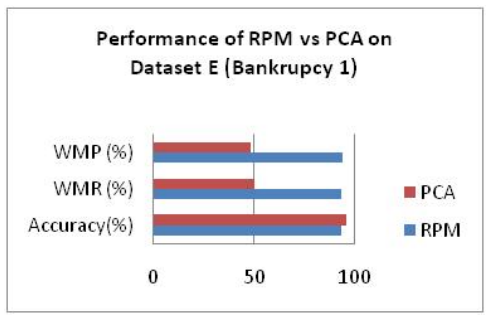

**Fig. 7: RPM and PCA for DATASET E (Bankruptcy 1)**

**Table 15: Results of Dataset E (Bankrupcy2) with PCA**

| Parameter | Result |
|---|---|
| Accuracy: | 96.05% +/- 0.11% |
| Kappa : | 0.017 +/- 0.031 |
| Weighted Mean Recall: | 50.47% +/- 0.83% |
| Weighted Mean Precision: | 60.55% +/- 20.19% |

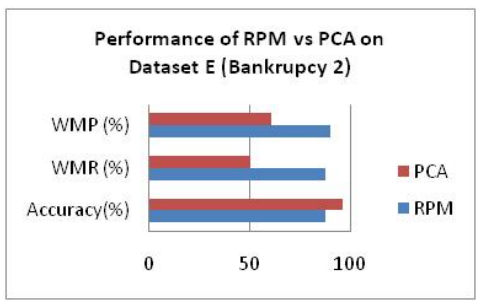

**Fig. 8: RPM and PCA for DATASET E (Bankruptcy 2)**

**Table 16: Results of Dataset E (Bankrupcy5) with PCA**

| Parameter | Result |
|---|---|
| Accuracy: | 93.13% +/- 0.11% |
| Kappa : | 0.026 +/- 0.028 |
| Weighted Mean Recall: | 50.71% +/- 0.80% |
| Weighted Mean Precision: | 69.91% +/- 23.85% |

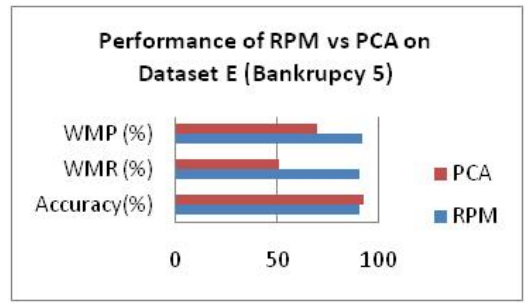

**Fig. 9: RPM and PCA for DATASET E (Bankruptcy 5)**

All the datasets worked well with RPM Model and attained accuracy level greater than 83% with Kappa Statistics greater than 0.6. Results showcased remarkable improvement in performance as compared to the performance with PCA. Pictorial representation of all datasets in terms of Accuracy Percentage and Kappa Statics is depicted below in Fig. 10 and 11.

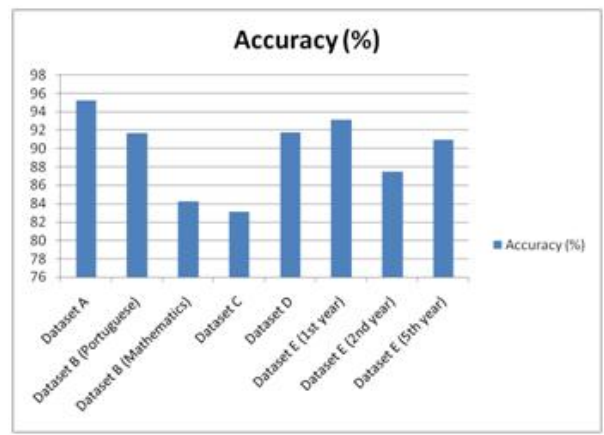

**Fig. 10: Prediction Accuracy of all Datasets (A-E)**

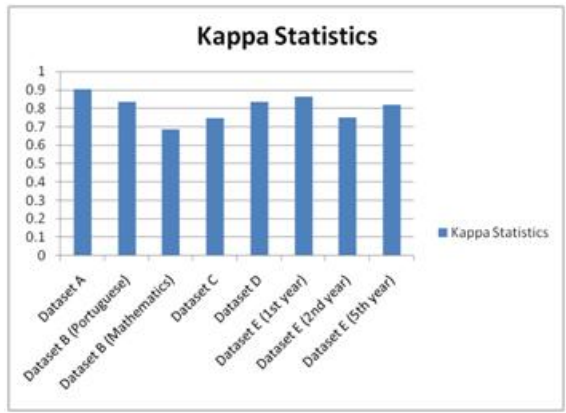

**Fig. 11: Kappa Statistics of All Datasets (A-E)**

## 5. Conclusion and Future Scope

The results prove that prediction performance can be enhanced by applying RPM when the dataset is large, unbalanced and multidimensional in nature. It is also proved that the model is robust in nature and can be applied to any type of dataset. The clustering applied to attributes set at pre-processing stage helps in parsimonious selection of variables and improves the performance of predictive algorithms. All the results showcase the robustness of RPM. This can be readily used by beginners for resolving real-world problems with easy application capabilities. In future RPM can be tested with big data.

**Acknowledgements**

The authors gratefully acknowledge the use of services and facilities of the Faculty Research Center and Library Resources at Vivekananda Institute of Professional Studies, GGSIPU, Delhi, India, where the first author is working as an Assistant Professor in Information Technology Department.